\documentclass[10pt,twocolumn,letterpaper]{article}

\usepackage{cvpr}              

\usepackage[accsupp]{axessibility}  

\usepackage{graphicx}
\usepackage{amsmath}
\usepackage{amssymb}
\usepackage{booktabs}
\usepackage{array}

\usepackage{bbding}
\usepackage{makecell}
\usepackage[normalem]{ulem}
\useunder{\uline}{\ul}{}
\usepackage[table,xcdraw]{xcolor}

\usepackage[pagebackref,breaklinks,colorlinks]{hyperref}

\usepackage[capitalize]{cleveref}
\crefname{section}{Sec.}{Secs.}
\Crefname{section}{Section}{Sections}
\Crefname{table}{Table}{Tables}
\crefname{table}{Tab.}{Tabs.}

\def\cvprPaperID{3094} 
\def\confName{CVPR}
\def\confYear{2022}

\begin{document}
	
	\title{MixSTE: Seq2seq Mixed Spatio-Temporal Encoder for 3D Human Pose Estimation in Video}
	
	\author{
    Jinlu Zhang$^{1}$ \qquad
    Zhigang Tu$^{1}$\thanks{Corresponding author: tuzhigang@whu.edu.cn}\qquad 
    Jianyu Yang$^{2}$\qquad
    Yujin Chen$^{3}$\thanks{Work done at Wuhan University}\qquad
    Junsong Yuan$^{4}$ \\
	$^{1}$Wuhan University\qquad
	$^{2}$Soochow University\qquad
	$^{3}$Technical University of Munich\qquad\\
	$^{4}$State University of New York at Buffalo\\
    {\tt\small \{jinluzhang, tuzhigang\}@whu.edu.cn,}
    {\tt\small jyyang@suda.edu.cn,}
    {\tt\small yujin.chen@tum.de,}
    {\tt\small jsyuan@buffalo.edu}
    }

	
	\maketitle
	
	\begin{abstract}
		Recent transformer-based solutions have been introduced to estimate 3D human pose from 2D keypoint sequence by considering body joints among all frames globally to learn spatio-temporal correlation.
		We observe that the motions of different joints differ significantly.
		However, the previous methods cannot efficiently model the solid inter-frame correspondence of each joint, leading to insufficient learning of spatial-temporal correlation.
		We propose MixSTE (Mixed Spatio-Temporal Encoder), which has a temporal transformer block to separately model the temporal motion of each joint and a spatial transformer block to learn inter-joint spatial correlation.
		These two blocks are utilized alternately to obtain better spatio-temporal feature encoding.
		In addition, the network output is extended from the central frame to entire frames of the input video, thereby improving the coherence between the input and output sequences.
		Extensive experiments are conducted on three benchmarks (\ie \textit{Human3.6M, MPI-INF-3DHP, and HumanEva}).
		The results show that our model outperforms the state-of-the-art approach by 10.9\% P-MPJPE and 7.6\% MPJPE. 
		The code is available at {\url{https://github.com/JinluZhang1126/MixSTE}}.

	\end{abstract}
	
	\begin{figure}[htp]
		\centering
		\includegraphics[width=1.0\linewidth]{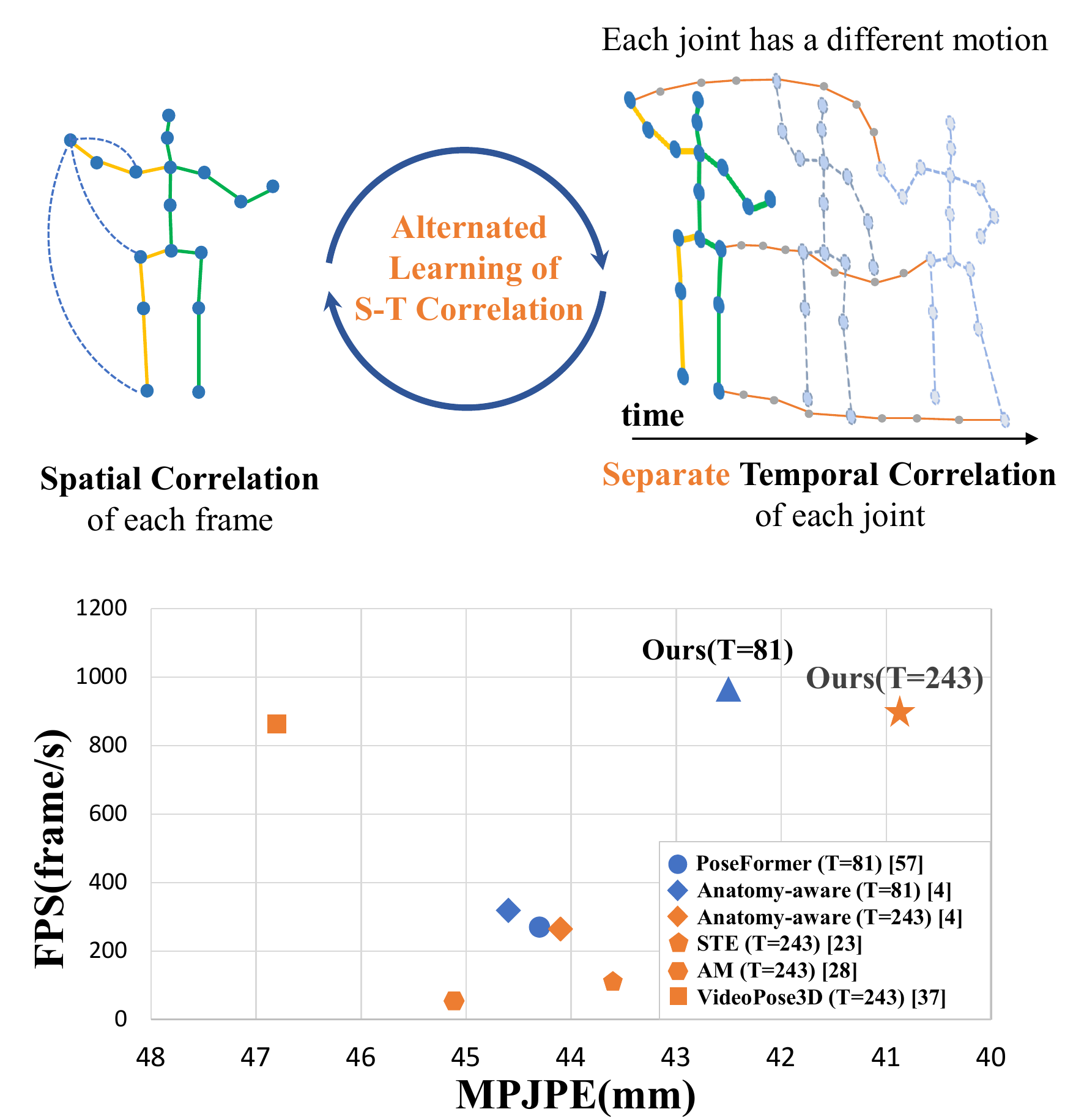}
		\vspace{-0.6cm}
		\caption{
			\textbf{Top:} Overview of spatio-temporal correlation modeling. Each 2D keypoint is separated in the temporal domain to learn different motion trajectories of body joints, and the spatial and temporal correlation are alternately stacked to improve the sequence coherence modeling ability.
			\textbf{Bottom:} Accuracy (MPJPE) and efficiency (FPS) comparison with different methods on Human3.6M dataset, the blue and orange colors indicate that the input sequence length $T$ is equal to 81 and 243, respectively.
		}
		\label{fig:01}
	\end{figure}
	
	\section{Introduction}
	\label{sec:intro}
	3D human pose estimation from monocular observations is a fundamental vision task that reconstructs 3D body joint locations from the input images or video.
	Since this task can obtain meaningful expressions of body geometry and motion, it has a wide range of applications, such as action recognition~\cite{weng2022action,zhang2020spatial}, virtual human~\cite{Yoon_2021_CVPR,chen2019so,chen2021joint, chen2021model}, and human-robot interaction~\cite{5152690,yemang,gong2022meta}.
	Most recent works are based on the 2D-to-3D lifting pipeline~\cite{simplebaseline,motionguidepose,cai2019exploiting,videopose3d,poseformer,attentionmechanism,anatomypose}, which detects 2D keypoints firstly and then lift them to 3D.
	Due to the depth ambiguity of monocular data, multiple potential 3D poses may be mapped from the same 2D pose, so it is difficult to recover an accurate 3D pose merely based on the information of a single frame 2D keypoints.
	
	Notable progress has been made by exploiting temporal information contained in the input video to address the above issues in a single frame~\cite{temporalconsis,videopose3d,attentionmechanism,anatomypose,motionguidepose,cai2019exploiting}.
	Recently, driven by the success of transformer~\cite{attisalluneed} for its ability to model sequence data, Zheng \etal~\cite{poseformer} introduces a transformer-based 3D human pose estimation network. 
	It takes advantage of spatio-temporal information for estimating the more accurate central-frame pose in video. 
	By modeling spatial correlations between all joints and temporal correlations among consecutive frames, PoseFormer~\cite{poseformer} achieves performance improvement.
	However, it ignores the motion differences among body joints, which causes the insufficient learning of spatio-temporal correlation.
	Moreover, it increases the dimension of the temporal transformer module, which limits the usage of longer input sequence.
	
	Poseformer \cite{poseformer} takes a video as input and only estimates the human pose of the central frame, which we summarize this pipeline as the \textit{seq2frame} approach.
	Many recent methods~\cite{videopose3d,attentionmechanism,anatomypose,cai2019exploiting,poseformer} follow it and they utilize adjacent frames to improve the accuracy of estimating the pose of a certain moment, but the sequence coherence is ignored due to the single frame output.
	Additionally, during the inference, these \textit{seq2frame} solutions need to input a 2D keypoint sequence repeatedly with large overlap to obtain 3D poses of all frames, which brings redundant calculation.
	In contrast to the \textit{seq2frame} approach, there is also the \textit{seq2seq} approach, which regresses the 3D pose sequence from the input 2D keypoints.
	These methods~\cite{motionguidepose,temporalconsis} mainly depend on long short-term memory (LSTM)~\cite{lstm} cell or graph convolution network (GCN)~\cite{gcn}, and perform well in learning temporal information among continuous estimation results.
	However, current \textit{seq2seq} networks lack the global modeling ability between input and output sequences, which tend to be excessively smooth~\cite{videopose3d} in the output poses of a long sequence.
	The low efficiency of LSTM~\cite{lstm} is also a severe issue for estimating human pose from video.
	
	While previous work has focused on associating all joints in the spatial and temporal domains, we observe that the motion trajectories of the different body joints vary from frame to frame and should be learned separately.
	Additionally, the input 2D keypoint sequence and the output 3D pose sequence have solid global coherence, and they should be tightly coupled to promote accurate and smooth 3D poses.
	
	Motivated by the above observations, in this work, we propose MixSTE to learn the separate temporal motion of each body joint and imbue sequential coherent human pose sequence in a \textit{seq2seq} approach.
	In contrast to the prior method~\cite{poseformer} which reconstructs the central frame and ignores the single joint motion, the MixSTE lifts 2D keypoint sequence to 3D pose sequence via a novel \textit{seq2seq} architecture and a set of motion-aware constraints.
	Specifically, as shown at the top of \Cref{fig:01}, we propose the joint separation to consider temporal motion information of each joint.
	It takes each 2D joint as an individual feature (which is referred to as a token in transformer) to sufficiently learn spatio-temporal correlation and helps to reduce the dimension of the joint features in temporal domain.
	Moreover, we propose an alternating design with \textit{seq2seq} to flexibly obtain better sequence coherence within a long sequence, which decreases redundant calculation and excessive smoothness.
	In this way, temporal motion trajectories of different body joints could be adequately considered to predict accurate 3D pose sequence.
	To the best of our knowledge, the proposed method is the first to utilize the transformer encoder in the \textit{seq2seq} pipeline, which enhances learning spatio-temperal correlation for accurate pose estimation and significantly improves the inference speed from seq2frame methods (see the bottom of Fig.1)
	Besides, our approach can easily adapt to any length of the input sequence. 
	
	
	Our contributions to 3D human pose estimation can be summarized in three folds:
	\begin{itemize}
		\item The MixSTE is proposed to effectively capture the temporal motion of different body joints over the long sequence, which helps to model sufficient spatio-temporal correlation.
		\item We propose  a novel alternating design with transformer-based \textit{seq2seq} model to learn the global coherence between sequences to improve the accuracy of reconstruction poses.
		\item Our approach achieves state-of-the-art performance on three benchmarks and has outstanding generalization.
	\end{itemize}
	
	
	\section{Related Work}
	\label{sec:rework}
	
	\textbf{3D Human Pose Estimation.}
	Estimating 3D human pose from monocular data was started by relying on the kinematics feature or the skeleton structure prior~\cite{ramakrishna_reconstructing_2012,ramakrishna2012reconstructing,ionescu_human36m_2014,ionescu_iterated_2014}.
	With the development of deep learning, more data-driven methods have been proposed, and these methods can be divided into end-to-end manner and 2D-to-3D lifting manner.
	The end-to-end manner directly estimates the 3D coordinates from the input without the intermediate 2D pose representation.
	Some methods~\cite{corsetofine,sun2018integral,tekin2016direct} followed this manner but required a high computation cost due to regressing directly from the image space.
	Different from the end-to-end manner, 2D-to-3D lifting pipeline first estimates 2D keypoints in the RGB data and then leverages the correspondences between 2D and 3D human structures to lift the 2D keypoints to 3D pose.
	Benefiting from the reliable effort of 2D keypoint detection works~\cite{he2017mask,chen2018cascaded,sun2019deep,newell2016stacked,ma2022remote}, recent 2D-to-3D lifting methods~\cite{semgcn,Ci_2019_ICCV,contextmodeling,graphstacked,liu2020comprehensive,simplebaseline,zhou2017towards} outperformed end-to-end approaches.
	Therefore, we follow the 2D-to-3D lifting manner to obtain robust 2D intermediate supervision.
	
	\textbf{\textit{Seq2frame} and \textit{Seq2seq} under 2D-to-3D Lifting.}
	Recently, temporal information from video has been exploited to produce more robust predictions by many methods.
	With the video input, many influential works (\textit{seq2frame}) pay attention to predicting the central frame of the input video to produce a more robust prediction and less sensitivity to noise.
	Pavllo \etal~\cite{videopose3d} proposed the dilated temporal convolutions based on the temporal convolution network (TCN) to extract temporal features.
	Some following works improved the performance of TCN by utilizing the attention mechanism~\cite{attentionmechanism}, or decomposing the pose estimation task into bone length and bone direction prediction~\cite{anatomypose}, but they have to fix the receptive field of the input sequence.
	In contrast to them, our approach is no need to preset the length of each input with respect to the convolution kernel or the sliding window size.
	Besides, GCN~\cite{gcn} was also applied to the task by~\cite{cai2019exploiting} to learn multi-scale features of human and hand poses.
	These works achieved good performance; however, calculation redundancy is a common flaw of these methods.
	
	On the other hand, some works (\textit{seq2seq}) improve the coherence and efficiency of 3D pose estimation and reconstruct all frames of input sequence at once.
	LSTM~\cite{lstm} was introduced to estimate 3D poses in video from a set of 2D keypoints~\cite{recurrent_pose_machine}. 
	Hossain \etal~\cite{temporalconsis} presented a temporal derivative loss function to ensure the temporal consistency over a sequence, but it faces the low computing efficiency issue.
	Wang \etal~\cite{motionguidepose} exploited a GCN-based approach and designed a corresponding loss to model motion in both short temporal intervals and long temporal ranges, but it lacks global modeling ability of input sequence.
	In contrast to~\cite{motionguidepose,temporalconsis}, our method has the advantage of global modeling ability of each joint in the spatial and temporal domains.
	Besides, it enables parallel processes for frames and joints to address the low-efficiency issue of LSTM~\cite{lstm}.
	
	\textbf{Self-attention and Transformer}
	The transformer architecture with self-attention was firstly proposed by~\cite{attisalluneed}, and then was applied to various visual tasks, \eg classification with visual transformer (ViT)~\cite{vit}, and detection with DETR~\cite{detr}.
	For the human pose estimation task,~\cite{yang2020transpose} proposed the Transpose to estimate 2D pose from images. 
	\cite{metro} presented a transformer framework for both human mesh recovery and pose estimation from a single image but ignored the temporal information in the video. 
	Some researchers also explored the multi-view 3D human pose estimation scheme~\cite{epipolar_transformer}.
	The stride transformer encoder~\cite{liftformer} was introduced to incorporate local contexts.
	Furthermore, PoseFormer~\cite{poseformer} constructed a model based on ViT~\cite{vit} to capture the spatial and temporal dependency sequentially.
	Both~\cite{liftformer} and~\cite{poseformer} have to fix the order of spatial and temporal encoders, and only the central frame of video is reconstructed.
	Our approach is similar to them in applying transformer architecture. 
	But we consider motion trajectories of different body joints and apply the \textit{seq2seq} to better model sequence coherence.
	
	From the above analysis and comparison of related works, further exploration for transformer-based methods in 3D human pose estimation is necessary and feasible, but there is no method combining the transformer with \textit{seq2seq} framework in the 3D human pose task.
	
	\begin{figure}[htp]
		\centering
		\includegraphics[width=1.0\linewidth]{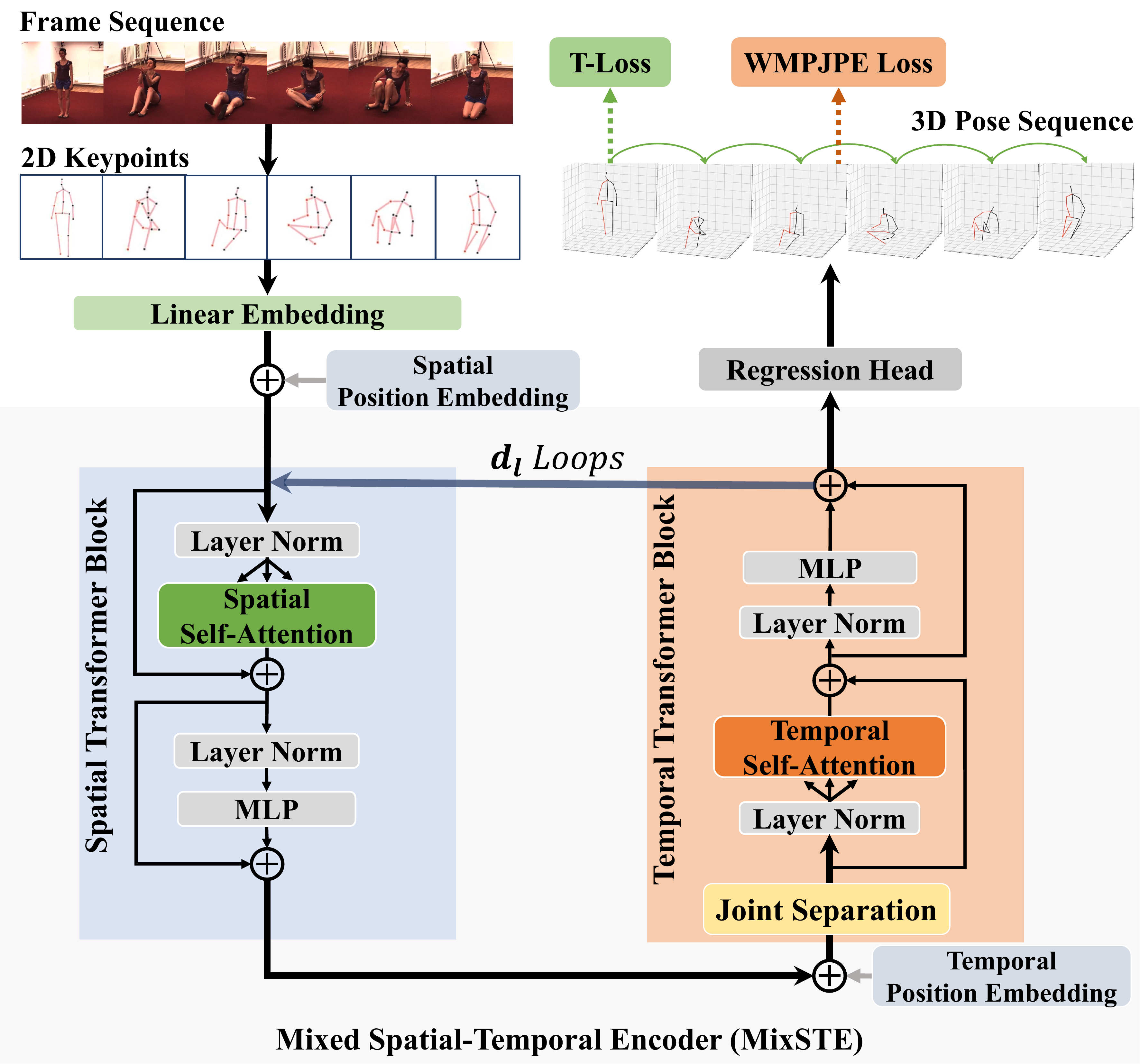}
		\caption{
			\textbf{Overview of the proposed framework.} 
			The MixSTE is stacked for $d_l$ loops, and each MixSTE models spatio-temporal dependencies independently. The WMPJPE Loss denotes the weighted per-joint position error loss. The T-Loss indicates the loss function of temporal coherence in \Cref{sec:multiloss}.
		}
		\label{fig:frame}
		\vspace{-0.3cm}
	\end{figure}
	
	\section{Our Approach}
	
	As shown in \Cref{fig:frame}, our network takes a concatenated 2D coordinates $C_{N,T}\in \mathbb{R}^{N\times T\times 2}$ with $N$ joints and $T$ frames as input, where the channel size of the input is 2.
	Firstly, we project the input keypoint sequence $C_{N,T}$ to high-dimensional feature $P_{N,T} \in \mathbb{R}^{N \times T \times d_m}$ with feature dimension $d_m$ for each joint representation.
	Then we utilize the position embedding matrix for retaining the position information of the spatial and temporal domains. 
	The proposed MixSTE takes the $P_{N,T}$ as input and aims to alternately learn the spatial correlation and separate temporal motion.
	Finally, we use a regression head to concatenate the outputs $X \in \mathbb{R}^{N\times T\times d_m}$ of encoder, and take the dimension $d_m$ to $3$ to get the 3D human pose sequence $Out \in \mathbb{R}^{N\times T\times 3}$. 
	
	\subsection{Mixed Spatio-Temporal Encoder}
	\label{sec:mixste}
	We utilize the MixSTE to model spatial dependency and temporal motion for a given 2D input keypoint sequence, respectively.
	MixSTE consists of a Spatial Transformer Block (STB) and a Temporal Transformer Block (TTB). 
	Here, the STB computes the self-attention between joints and aims to learn the body joint relations of each frame, while the TTB computes the self-attention between frames and focuses on learning the global temporal correlation of each joint.
	\subsubsection{Separate Temporal Correlation Learning}
	To imbue effective motion trajectories into the learned representations, we consider the temporal correspondence of each joint in order to explicitly model correlations on the same joint over the dynamic sequence.
	Different from the previous method~\cite{poseformer}, we do not treat all body joints as a token in the temporal transformer block.
	We separate different joints in time dimension, so that the trajectory of each joint is an individual token $p \in \mathbb{R}^{1 \times T \times d_m}$, and different joints of body are modeled paralleled.
	From the perspective of the time dimension, different motion trajectories of body joints are modeled separately to represent temporal correlations better.
	The joint separation is operated as follows:
	\begin{equation}
		\begin{aligned}
			X_l^{t} = Concat(\mathcal{F}(p_{i,1},p_{i,2},...p_{i,T})), \ i\in N,
		\end{aligned}
	\end{equation}
	where $p_{i,j} \in P_{N,T}$ denotes the $i$-th joint in the $j$-th frame, $\mathcal{F}$ indicates the temporal encoder function and the output of the $l$-th TTB encoder is $X_l \in \mathbb{R}^{N \times T \times d_m}$.
	Furthermore, treating each body joint as an individual token can decrease dimension of the model to $d_m$ from $N \times d_m$ of PoseFormer~\cite{poseformer}, and it also enables the longer sequence processed in the model.
	
	
	\subsubsection{Spatial Correlation Learning}
	We employ the spatial transformer block (STB) to learn spatial correlations among joints in each frame.
	Given 2D keypoints with $N$ joints, we consider each joint as a token in spatial attention.
	Firstly, we take 2D keypoints as input and project each keypoint to a high-dimensional feature with the linear embedding layer. 
	The feature is referred to as a spatial token in STB.
	We then embed the spatial position information with a positional matrix $E_{s-pos} \in \mathbb{R}^{N \times d_m}$.
	After that, spatial tokens $P_i \in \mathbb{R}^{N\times d_m}$ of the $i$-th frame is fed into spatial self-attention mechanism of STB to model dependencies across all joints and output the high-dimensional tokens $X_l^{s} \in \mathbb{R}^{N \times T \times d_m}$ in $l$-th STB.
	
	\subsubsection{Alternating design with \textit{Seq2seq}}
	\textbf{Alternating design in spatio-temporal correlation.}
	The STB and TTB are designed in an alternating way to encode different high-dimensional tokens.
	The process of alternating design is like recurrent neural network~(RNN), but we can parallel over joint and time dimensions.
	We stack STB and TTB for $d_l$ loops, and the dimension of the feature is preserved as a fixed size $d_m$ to promise that spatial-temporal correlation learning focuses on the same joint.
	Specifically, the spatial and temporal position embedding is applied only in the first encoder to retain two kinds of position information. 
	Moreover, there is the independence of the spatial and temporal domains, where previous methods often only learn partial sequence coherence due to the single process of spatio-temporal modeling.
	The proposed alternating design with stacking architecture can obtain better coherence and spatio-temporal feature encoding.
	
	\noindent \textbf{\textit{Seq2seq} framework.}
	Furthermore, to better utilize the global sequence coherence between the input sequence of 2D keypoints and the output sequence of 3D poses, we leverage the \textit{seq2seq} pipeline in our model.
	It can predict all 3D poses of input 2D keypoints at once, which helps to preserve sequence coherence between the input and output sequences.
	Besides, for a sequence containing $T$ frames, we need fewer times of inference, which means higher efficiency.
	Assuming that the sequence length of each input $t<T$, the inference time gap $G$ between our model and the \textit{seq2frame} methods will become higher with the increase of $t$:
	\begin{equation}
		G = \frac{T(1+2\delta)}{(\frac{T+2\delta}{t})} = \frac{T(1+2\delta)}{T + 2 \cdot \delta} \cdot t \approx (1+2\delta) \cdot t,
		\label{eq:seq2seq_inf_times}
	\end{equation}
	where $\delta$ indicates the padding length of the input sequence.
	
	In summary, due to these advanced components, our model can capture various temporal motions and global sequence coherence with less calculation redundancy.
	
	\subsection{Transformer Block in MixSTE}
	\label{sec:stb&ttb}
	The transformer blocks in MixSTE follow the scaled dot-product attention~\cite{attisalluneed}. The attention computing of query, key, and value matrix $Q, K, V$ in each head are formulated by:
	\begin{equation}
		\begin{aligned}
			Attention(Q,K,V)=Softmax(\frac{QK^T}{\sqrt{d_m}})V,
		\end{aligned}
	\end{equation}
	where $\{Q,K,V\} \in \mathbb{R}^{N \times d_m}$, $N$ indicates the number of tokens, and $d_m$ is the dimension of each token.
	The concatenated attention of $h$ heads is defined as follows:
	\begin{equation}
		\begin{aligned}
			MSA = Concat(head1,..., head_h)W^{O},
		\end{aligned}
	\end{equation}
	\begin{equation}
		head_i = Attention(Q_i,K_i,V_i), i \in h,
	\end{equation}
	where the linear projection weight is $W^{O} \in \mathbb{R}^{d_m \times d_m} $.
	In the transformer encoder of our approach, each joint token $p \in P_{N}$ is projected from joint $c_i$ of the 2D coordinates $C_{N} \in \mathbb{R}^{N \times 2}$. 
	Joint token $p$ is embedded with the position information by a matrix $E_{pos} \in \mathbb{R}^{N \times d_m}$:
	\begin{equation}
		X = Norm(L_e(c_i)+E_{pos}), \ X \in \mathbb{R}^{N\times d_m},
	\end{equation}
	where $Norm$ denotes the layer normalization, and $L_e$ indicates the linear embedding layer.
	The spatial-temporal dependencies among joints are then computed by the STB and TTB as follows:
	\begin{equation}
		R_s = MSA(U_Q,U_K,U_V) + X,
	\end{equation}
	\begin{equation}
		U_i = XW^{m}, \ m \in \{Q,K,V\},
	\end{equation}
	where $R_s$ denotes the attention output of the joint token $X$, $U_i$ is the matrix mapped from $X$ by linear transformation, and $W^m$ is the corresponding linear transformation weight matrix of query, key and value in joints.

	\subsection{Loss Function}
	\label{sec:multiloss}
	The network is trained in an end-to-end manner, the final loss function $\mathcal{L}$ is defined as:
	\begin{equation}
		\mathcal{L} = \mathcal{L}_{w} + \lambda_t \mathcal{L}_t + \lambda_m \mathcal{L}_m,
	\end{equation}
	where $\mathcal{L}_{w}$ is the WMPJPE loss, $\mathcal{L}_t$ is the TCLoss, and $\mathcal{L}_m$ denotes the MPJVE loss. 
	During the training stage, different coefficients $\lambda_t$ and $\lambda_m$ are employed to $\mathcal{L}_t$ and $\mathcal{L}_m$ to avoid excessive smoothness in sequence. 
	
	In detail, we firstly explored a weighted mean per-joint position error (WMPJPE), which pays different attention to different joints of the human body when computing the MPJPE.
	The WMPJPE $L_{w}$ with weight $W$ is computed as follows:
	\begin{equation}
		\mathcal{L}_{w} = \frac{1}{N^s}\sum^{N^s}_{i=1}(W \times \frac{1}{T}\sum^{T}_{j=1}\parallel p_{i,j} - gt_{i,j}\parallel^2_2)),
		\label{eq:wmpjpe}
	\end{equation}
	where $N^s$ indicates $N$ joints of human skeleton $s$ in three datasets, $T$ denotes the number of frames in sequence, $p_{i,j}$ and $gt_{i,j}$ are the prediction and the ground truth 3D pose of $i$-th joint in $j$-th frame.
	
	Moreover, the temporal consistency loss (TCLoss) in~\cite{temporalconsis} is introduced to produce the smooth poses.
	The MPJVE~\cite{videopose3d} is also a loss in our model to improve the temporal coherence between the predicted pose sequence and the ground truth sequence.
	We merge the TCLoss and MPJVE as the temporal loss function (T-Loss).
	
	\section{Experiment}
	\begin{table*}[htp]
		\centering
		\resizebox{\textwidth}{!}{%
			\begin{tabular}{l@{ }l|ccccccccccccccc|c}
				\toprule
				\textbf{Protocol \#1}  &      & Dir.          & Disc.         & Eat           & Greet         & Phone         & Photo         & Pose          & Pur.          & Sit           & SitD.         & Smoke         & Wait          & WalkD.        & Walk          & WalkT.        & \textbf{Avg.}  \\
				\hline
				Pavlakos \etal~\cite{ordinaldepth}                          & CVPR2018      & 48.5          & 54.4          & 54.4          & 52.0          & 59.4          & 65.3          & 49.9          & 52.9          & 65.8          & 71.1          & 56.6          & 52.9          & 60.9          & 44.7          & 47.8          & 56.2          \\
				Pavllo \etal~\cite{videopose3d}(CPN, $T$=243)(†)            & CVPR2019   & 45.2          & 46.7          & 43.3          & 45.6          & 48.1          & 55.1          & 44.6          & 44.3          & 57.3          & 65.8          & 47.1          & 44.0          & 49.0          & 32.8          & 33.9          & 46.8          \\
				Cai \etal~\cite{cai2019exploiting}(CPN, $T$=7)(†)        				  & ICCV2019      & 44.6          & 47.4          & 45.6          & 48.8          & 50.8          & 59.0          & 47.2          & 43.9          & 57.9          & 61.9          & 49.7          & 46.6          & 51.3          & 37.1          & 39.4          & 48.8          \\
				Yeh \etal~\cite{chiralitynets}(†)                         & NIPS2019      & 44.8          & 46.1          & 43.3          & 46.4          & 49.0          & 55.2          & 44.6          & 44.0          & 58.3          & 62.7          & 47.1          & 43.9          & 48.6          & 32.7          & 33.3          & 46.7          \\
				Liu \etal~\cite{attentionmechanism}(CPN, $T$=243)(†)   & CVPR2020      & 41.8          & 44.8          & 41.1          & 44.9          & 47.4          & 54.1          & 43.4          & 42.2          & 56.2          & 63.6          & 45.3          & 43.5          & 45.3          & 31.3          & 32.2          & 45.1          \\
				Wang \etal~\cite{motionguidepose}(CPN, $T$=96)(†)                   & ECCV2020      & 40.2          & 42.5          & 42.6          & 41.1          & 46.7          & 56.7          & 41.4          & 42.3          & 56.2          & 60.4          & 46.3          & 42.2          & 46.2          & 31.7          & 31.0          & 44.5          \\
				Chen \etal~\cite{anatomypose}(CPN, $T$=243)(†)             & TCSVT2021 & 41.4          & 43.5          & 40.1          & 42.9          & 46.6          & 51.9          & 41.7          & 42.3          & 53.9          & 60.2          & 45.4          & 41.7          & 46.0          & 31.5          & 32.7          & 44.1          \\
				Xu \etal~\cite{graphstacked}($T$=1)         & CVPR2021      & 45.2          & 49.9          & 47.5          & 50.9          & 54.9          & 66.1          & 48.5          & 46.3          & 59.7          & 71.5          & 51.4          & 48.6          & 53.9          & 39.9          & 44.1          & 51.9          \\
				Lin \etal~\cite{metro}($T$=1)(*)                        & CVPR2021      & -             & -             & -             & -             & -             & -             & -             & -             & -             & -             & -             & -             & -             & -             & -             & 54.0          \\
				Zeng \etal~\cite{hardpose}(†) & ICCV2021	& 43.1 & 50.4 & 43.9 & 45.3 & 46.1 & 57.0 & 46.3 & 47.6 & 56.3 & 61.5 & 47.7 & 47.4 & 53.5 & 35.4 & 37.3 & 47.9	\\
				Zheng \etal~\cite{poseformer}(CPN, $T$=81)(†)(*)          & ICCV2021      & 41.5          & 44.8          & 39.8          & 42.5          & 46.5          & 51.6          & 42.1          & 42.0          & 53.3          & 60.7          & 45.5          & 43.3          & 46.1          & 31.8          & 32.2          & 44.3          \\
				\rowcolor[HTML]{DADADA} 
				Ours(CPN, $T$=81)(†)(*)				  &               & \uline{39.8}        & \uline{43.0}          & {\ul 38.6}    & \uline{40.1}        & {\ul 43.4}    & \uline{50.6}          & \uline{40.6}          & {\ul 41.4}    & \uline{52.2}    & {\ul 56.7}    & {\ul 43.8}    & \uline{40.8}          & {\ul 43.9}    & {\ul 29.4}    & {\ul 30.3}    & {\ul 42.4}    \\
				\rowcolor[HTML]{DADADA}
				Ours(CPN, $T$=243)(†)(*)      &               & \textbf{37.6}    & \textbf{40.9} & \textbf{37.3} & \textbf{39.7} & \textbf{42.3} & \textbf{49.9}    & \textbf{40.1}    & \textbf{39.8} & \textbf{51.7}          & \textbf{55.0} & \textbf{42.1} & \textbf{39.8} & \textbf{41.0} & \textbf{27.9} & \textbf{27.9} & \textbf{40.9} \\
				
				\hline
				Wang \etal~\cite{motionguidepose}(HRNet, $T$=96)(†) 	& ECCV2020      & {\ul 38.2}    & {\ul 41.0}    & 45.9          & {\ul 39.7}    & {\ul 41.4}    & {\ul 51.4}    & 41.6          & 41.4          & 52.0          & 57.4    & {\ul 41.8}    & {\ul 44.4}    & {\ul 41.6}    & {\ul 33.1}    & \textbf{30.0} & {\ul 42.6}    \\
				Wehrbein \etal~\cite{probabilistic}(HRNet, $T$=200)          & ICCV2021      & 38.5          & 42.5          & {\ul 39.9}    & 41.7          & 46.5          & 51.6          & {\ul 39.9}    & {\ul 40.8}    & {\ul 49.5}    & \uline{56.8} & 45.3          & 46.4          & 46.8          & 37.8          & 40.4          & 44.3          \\
				\rowcolor[HTML]{DADADA}
				Ours(HRNet, $T$=243)            &               & \textbf{36.7} & \textbf{39.0} & \textbf{36.5} & \textbf{39.4} & \textbf{40.2} & \textbf{44.9} & \textbf{39.8} & \textbf{36.9} & \textbf{47.9} & \textbf{54.8}          & \textbf{39.6} & \textbf{37.8} & \textbf{39.3} & \textbf{29.7} & {\ul 30.6}    & \textbf{39.8}      \\
				\hline
				\midrule
				\textbf{Protocol \#2}            &             & Dir.          & Disc.         & Eat           & Greet         & Phone         & Photo         & Pose          & Pur.          & Sit           & SitD.         & Smoke         & Wait          & WalkD.        & Walk          & WalkT.        & \textbf{Avg.}  \\
				\hline
				Wang \etal~\cite{motionguidepose}(CPN, $T$=96)(†)          	& ECCV2020    & \uline{31.8}          & 34.3          & 35.4          & \uline{33.5}          & 35.4          & 41.7          & \textbf{31.1}          & {\ul 31.6}    & 44.4          & 49.0          & 36.4          & 32.2          & 35.0          & 24.9          & \uline{23.0} & 34.5          \\
				Liu \etal~\cite{attentionmechanism}(CPN, $T$=243)(†) 		& CVPR2020    & 32.3          & 35.2          & 33.3          & 35.8          & 35.9          & 41.5          & 33.2          & 32.7          & 44.6          & 50.9          & 37.0          & 32.4          & 37.0          & 25.2          & 27.2          & 35.6          \\
				
				Zheng \etal~\cite{poseformer}(CPN, $T$=81)(†)(*)        		& ICCV2021    & 34.1          & 36.1          & 34.4          & 37.2          & 36.4          & 42.2          & 34.4          & 33.6          & 45.0          & 52.5          & 37.4          & 33.8          & 37.8          & 25.6          & 27.3          & 36.5          \\
				\rowcolor[HTML]{DADADA}
				Ours(CPN, $T$=81)(†)(*)     &             & 32.0          & {\ul 34.2}    & {\ul 31.7}    & 33.7          & {\ul 34.4}    & \uline{39.2}          & 32.0          & 31.8          & 42.9          & {\ul 46.9}    & \uline{35.5}          & \uline{32.0}          & {\ul 34.4}    & {\ul 23.6}    & 25.2          & {\ul 33.9}    \\
				\rowcolor[HTML]{DADADA}
				Ours(CPN, $T$=243)(†)(*)    &             & \textbf{30.8}    & \textbf{33.1} & \textbf{30.3} & \textbf{31.8} & \textbf{33.1} & \textbf{39.1}    & \textbf{31.1}    & \textbf{30.5} & \textbf{42.5}    & \textbf{44.5} & \textbf{34.0} & \textbf{30.8} & \textbf{32.7} & \textbf{22.1} & \textbf{22.9}    & \textbf{32.6} \\
				\hline
				Wang \etal~\cite{motionguidepose}(HRNet)(†)                        & ECCV2020    & 28.4          & 32.5          & 34.4          & 32.3          & {\ul 32.5}    & 40.9          & 30.4          & 29.3          & 42.6          & \textbf{45.2} & {\ul 33.0}    & {\ul 32.0}    & {\ul 33.2}    & {\ul 24.2}    & {\ul 22.9}    & 32.7          \\
				Wehrbein \etal~\cite{probabilistic}(HRNet, $T$=200)        & ICCV2021    & \textbf{27.9} & {\ul 31.4}    & {\ul 29.7}    & \textbf{30.2} & 34.9          & {\ul 37.1}    & \textbf{27.3} & {\ul 28.2}    & {\ul 39.0}    & {\ul 46.1}    & 34.2          & 32.3          & 33.6          & 26.1          & 27.5          & {\ul 32.4}    \\
				\rowcolor[HTML]{DADADA}
				Ours(HRNet, $T$=243)        &             & {\ul 28.0}    & \textbf{30.9} & \textbf{28.6} & {\ul 30.7}    & \textbf{30.4} & \textbf{34.6} & {\ul 28.6}    & \textbf{28.1} & \textbf{37.1} & 47.3          & \textbf{30.5} & \textbf{29.7} & \textbf{30.5} & \textbf{21.6} & \textbf{20.0} & \textbf{30.6} \\
				\midrule
				\hline
				\textbf{MPJVE}            &             & Dir.          & Disc.         & Eat           & Greet         & Phone         & Photo         & Pose          & Pur.          & Sit           & SitD.         & Smoke         & Wait          & WalkD.        & Walk          & WalkT.        & \textbf{Avg.}  \\
				Pavllo \etal~\cite{videopose3d}(†)       & CVPR2019	 & 3.0      & 3.1       & 2.2       & 3.4       & 2.3       & 2.7           & 2.7       & 3.1       & 2.1       & 2.9       & 2.3       & 2.4       & 3.7       & 3.1       & 2.8       & 2.8		\\
				Chen \etal~\cite{anatomypose}(†)       & TCSVT2021       & 2.7        & 2.8        & \textbf{2.0}        & 3.1        & 2.0        & 2.4        & 2.4        & 2.8        & 1.8        & 2.4        & 2.0        & \textbf{2.1}        & 3.4        & \textbf{2.7}        & 2.4        & 2.5 		\\
				Zheng \etal~\cite{poseformer}(†)(*)       & ICCV2021       & 3.2       & 3.4       & 2.6       & 3.6       & 2.6       & 3.0       & 2.9       & 3.2       & 2.6       & 3.3       & 2.7       & 2.7       & 3.8       & 3.2       & 2.9       & 3.1		\\
				\rowcolor[HTML]{DADADA}
				Ours(CPN, $T$=243)(†)(*) 				  &        & \textbf{2.5}       & \textbf{2.7}       & \textbf{1.9}       & \textbf{2.8}       & \textbf{1.9}       & \textbf{2.2}       &\textbf{2.3}		  & \textbf{2.6}       & \textbf{1.6}       & \textbf{2.2}       & \textbf{1.9}       & \textbf{2.0}       & \textbf{3.1}       & \textbf{2.6}       & \textbf{2.2}       & \textbf{2.3}		\\
				\bottomrule
			\end{tabular}%
		}
		\vspace{-0.3cm}	
		\caption{Detailed quantitative comparison results of MPJPE in millimeters (mm) on Human3.6M under Protocol 1 (no rigid alignment applied) and Protocol 2 (rigid alignment). \textbf{Top table}: results under Protocol 1 (MPJPE); \textbf{Middle table}: results under Protocol 2 (P-MPJPE); \textbf{Bottom table}: results of MPJVE. $T$ denotes the number of input frames estimated by the respective approaches, (†) indicates using temporal information, and (*) indicates the transformer-based methods. The best and second-best results are highlighted in bold and underlined formats, respectively.}
		\label{tab:h36m_cpn_hr}
	\end{table*}
	\begin{table*}[htp]
		\centering
		\resizebox{\textwidth}{!}{%
			\begin{tabular}{l@{}l|ccccccccccccccc|c}
				\toprule
				\textbf{Protocol   \#1}                    &               & Dir. & Disc. & Eat  & Greet & Phone & Photo & Pose & Pur. & Sit  & SitD. & Smoke & Wait & WalkD. & Walk & WalkT. & \textbf{Avg.}  \\
				\midrule
				Liu \etal~\cite{attentionmechanism}($T$=243)(†) & CVPR2020      & 34.5 & 37.1  & 33.6 & 34.2  & 32.9  & 37.1  & 39.6 & 35.8 & 40.7 & 41.4  & 33.0  & 33.8 & 33.0   & 26.6 & 26.9   & 34.7 \\
				Wang \etal~\cite{motionguidepose}(GT, $T$=96)                    & ECCV2020      & 23.0 & 25.7  & 22.8 & 22.6  & 24.1  & 30.6  & 24.9 & 24.5 & 31.1 & 35.0  & 25.6  & 24.3 & 25.1   & 19.8 & 18.4   & 25.6 \\
				Zheng \etal~\cite{poseformer}(T = 81)(†)(*)     & ICCV2021      & 30.0 & 33.6  & 29.9 & 31.0  & 30.2  & 33.3  & 34.8 & 31.4 & 37.8 & 38.6  & 31.7  & 31.5 & 29.0   & 23.3 & 23.1   & 31.3 \\
				\rowcolor[HTML]{DADADA}
				Ours($T$=81)                        &               & 25.6 & 27.8  & 24.5 & 25.7  & 24.9  & 29.9  & 28.6 & 27.4 & 29.9 & 29.0  & 26.1  & 25.0 & 25.2   & 18.7 & 19.9   & 25.9 \\
				\rowcolor[HTML]{DADADA}
				Ours($T$=243)                       &               & \textbf{21.6} & \textbf{22.0} & \textbf{20.4} & \textbf{21.0} & \textbf{20.8} & \textbf{24.3} & \textbf{24.7} & \textbf{21.9} & \textbf{26.9} & \textbf{24.9} & \textbf{21.2} & \textbf{21.5} & \textbf{20.8} & \textbf{14.7} & \textbf{15.7} & \textbf{21.6}\\
				\bottomrule
			\end{tabular}
		}
		\vspace{-0.3cm}
		\caption{Detailed quantitative comparison results of MPJPE in millimeters (mm) on Human3.6M under Protocol 1 using 2D ground truth keypoints as input. The best results are highlighted in bold.}
		\label{tab:h36m_gt}
	\end{table*}
	
	\subsection{Datasets and Evaluation Protocols}
	We evaluate our model on three 3D human pose estimation datasets: Human3.6M~\cite{h36m_iccv, h36m_pami}, MPI-INF-3DHP~\cite{3dhp} and HumanEva~\cite{heva} individually. 
	
	\textbf{Human3.6M} is the most commonly used indoor dataset for the 3D human pose estimation tasks. 
	Following the same policy of previous methods~\cite{corsetofine, simplebaseline, ordinaldepth, videopose3d, attentionmechanism, poseformer, anatomypose}, the 3D human pose in Human3.6M is adopted as a 17-joint skeleton, and the subjects \textit{S1, S5, S6, S7, S8} from the dataset are applied during training, the subjects \textit{S9} and \textit{S11} are used for testing. 
	The two commonly used evaluation metrics (MPJPE and P-MPJPE) are involved in this dataset. 
	In addition, mean per-joint velocity error (MPJVE)~\cite{videopose3d} is applied to measure the smoothness of the prediction sequence. 
	We also compute the variance (VAR.) of MPJPE between action categories to evaluate the stability.
	
	\textbf{MPI-INF-3DHP} is also a recently popular large-scale 3D human pose dataset. 
	Our setting follows previous works~\cite{motionguidepose,poseformer}.
	The area under the curve (AUC), percentage of correct keypoints (PCK), and MPJPE are reported as evaluation metrics.
	
	\textbf{HumanEva} is a smaller dataset than above datasets.
	As the same setting of~\cite{attentionmechanism, poseformer}, actions (Walk, Jog) in subjects \textit{S1, S2, S3} are evaluation data. 
	The metrics MPJPE and P-MPJPE are applied.
	
	\subsection{Implementation Details}
	The proposed model is implemented with Pytorch.
	We use 2D keypoints from 2D pose detector~\cite{sun2019deep, chen2018cascaded} or 2D ground truth to analyze the performance of our framework.
	Although the proposed model can easily adapt to any length of input sequence, to be fair, we select some specific sequence lengths $T$ for three datasets to compare our method with other methods which must have a certain 2D input length~\cite{videopose3d,anatomypose,attentionmechanism}: Human3.6M ($T$=81,243), MPI-INF-3DHP ($T$=1,27), HumanEva ($T$=43).
	Analysis about the frame length setting is discussed in the ablation study \Cref{sec:ablation}.
	The $W$ in WMPJPE is set
    based on different joint groups (torso, head, middle limb,
    and terminal limb) with different values (1.0, 1.5, 2.5, and
    4.0, respectively).
	The Adam optimizer~\cite{kingma2014adam} is employed for the training model.
	The batch size, dropout rate, and activation function for datasets are set to 1024, 0.1, and GELU.
	We utilize the stride data sample strategy with interval is as same as the input length to make there no overlapping frames between sequences(more details in the supplementary material).
	
	\subsection{Comparison with State-of-the-art Methods}
	\textbf{Results on Human3.6M.} 
	Two types of 2D joint detection data are applied in the experiment: CPN~\cite{chen2018cascaded}, which is the most typical 2D estimator used in previous approaches, and HRNet~\cite{sun2019deep} which is used to further investigate the upper bound of our method. 
	The results compared with other methods, including the error of all 15 actions and the average error, are reported in \Cref{tab:h36m_cpn_hr}.
	For CPN~\cite{chen2018cascaded} detector, our model obtains the best result of average MPJPE of 40.9mm under Protocol 1 and 32.6mm P-MPJPE under Protocol 2, which outperforms PoseFormer~\cite{poseformer} by 3.4mm MPJPE (\textbf{7.6\%}).
	Furthermore, our method achieves the best under $T=243$ setting and second-best under $T=81$ setting in all actions.
	
	Utilizing more powerful 2D detector HRNet~\cite{sun2019deep}, our model further improves roughly 4.5mm (\textbf{10.2\%}) under Protocol 1.
	We also compare our method with~\cite{videopose3d,poseformer,attentionmechanism,anatomypose,motionguidepose} using 2D ground truth, and the results are illustrated in the \Cref{tab:h36m_gt}.
	Our method significantly outperforms all other methods and achieves approximately \textbf{31.0\%} improvement of average MPJPE compared with PoseFormer~\cite{poseformer}.
	
\begin{figure}[htp]
		\centering
		\includegraphics[width=1.0\linewidth]{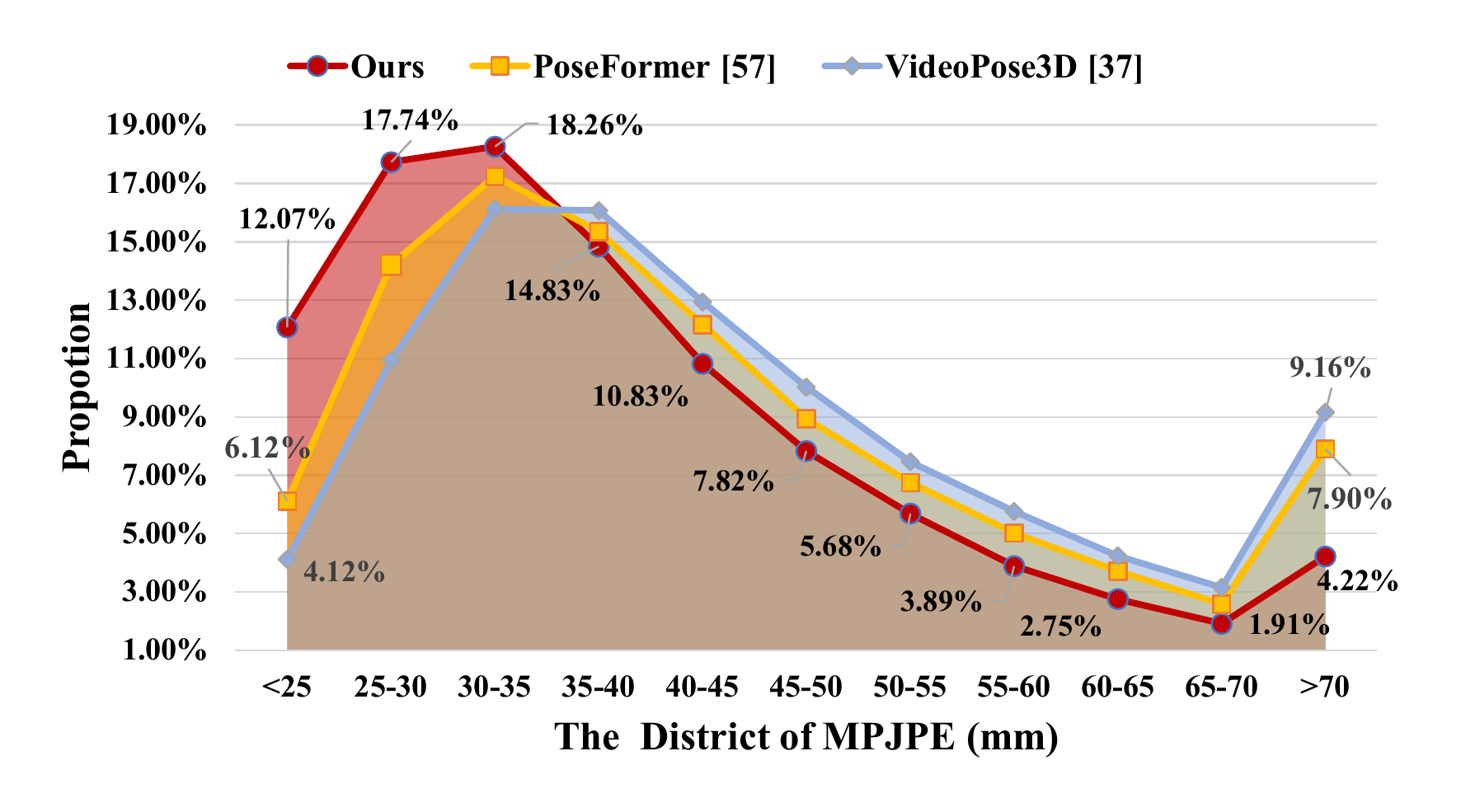}
		\vspace{-0.8cm}
		\caption{The MPJPE distribution on Human3.6M testset.}
		\label{fig:proportion}
		\vspace{-0.3cm}
	\end{figure}

	
	Furthermore, we compare	the MPJPE distribution in the testset \textit{S9} and \textit{S11} with other methods~\cite{poseformer,videopose3d} to evaluate the ability of estimating difficult poses. 
	It can be observed in \Cref{fig:proportion} that there are much	fewer poses with high errors in our method.
	Moreover, the proportion of poses with over 40mm MPJPE, which causes loss of accuracy, is consistently lower, and the proportion of less than 30mm MPJPE is much higher than other methods.
	The results demonstrate our method performs better on difficult actions.
	
	\begin{figure}[htp]
		\centering
		\includegraphics[width=1.0\linewidth]{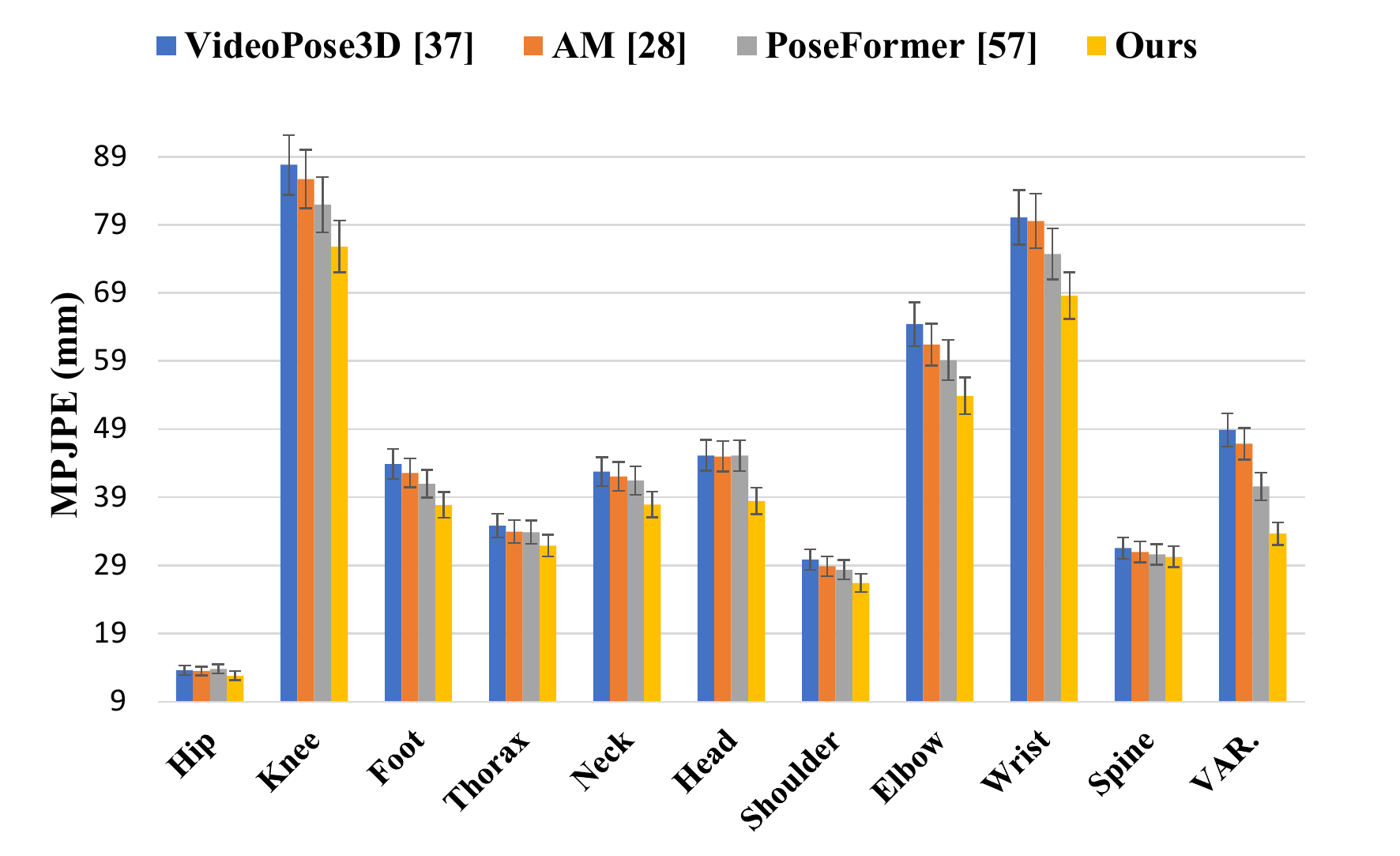}
		\vspace{-0.8cm}
		\caption{
			The average joint error comparison across all frames of the testset in the Human3.6M. 
			The $VAR.$ indicates the variance among joint errors divided by a factor (10.0), and the joints of the same part (\eg right knee and left knee) are divided into the same category for the sake of display.
		}
		\label{fig:jointserror}
	\end{figure}
	
	In \Cref{fig:jointserror}, we compare the MPJPE for individual joints on all frames of Human3.6M testset to evaluate the estimation accuracy of different joints.
	The joints of limbs have higher errors due to flexible movements, while the trunk joints have lower errors because of stable motion.
	Our accuracy of each joint category achieves the best, and the variance ($VAR.$) comparison shows that our method has a more stable performance.
	
	\textbf{Results on MPI-INF-3DHP.} 
	\Cref{tab:3dhp} reports the detailed comparison with other methods on the MPI-INF-3DHP testset.
	In addition, the 1-frame setting is employed to evaluate the single-frame performance.
	The input is ground truth 2D keypoints.
	As shown in the table, the method ($T$=27) performs the best in three evaluation metrics, and the single-frame setting ($T$=1) also achieves the second-best accuracy. These results demonstrate the strong performance of our model in single-frame and multi-frame scenarios.
	
	\textbf{Reults on HumanEva.} 
	
	We utilize HumanEva to evaluate the generalization ability of the proposed method and the impact of finetuning from large datasets.
	The MPJPE results on HumanEva finetuning from Human3.6M are reported in the \Cref{tab:heva_ft}.
	Due to \textit{seq2seq} setting and limitation of transformer in small dataset, our method without fine-tuning is slightly worse than our baseline.
	But the performance can be improved by using smaller data sample strides (interval=1).
	The experiment shows that our model has a better generalization ability than previous methods.
	\begin{table}[tp]
		\centering
		\resizebox{230pt}{!}{
			\begin{tabular}{l@{}l|ccc}
				\toprule
				Method                               &                & PCK$\uparrow$          & AUC$\uparrow$          & MPJPE$\downarrow$        \\
				\midrule
				Mehta \etal~\cite{vnect2017mehta}               & ACM TOG 2017   & 79.4          & 41.6          & -             \\
				Lin \etal~\cite{lin2019trajectory}($T$=25) 	 & BMVC2019       & 83.6          & 51.4          & 79.8          \\
				Li \etal~\cite{evolutionary_Li_2020_CVPR}    & CVPR2020       & 81.2          & 46.1          & 99.7          \\
				Wang \etal~\cite{motionguidepose}($T$=96)        & ECCV2020       & 86.9          & 62.1          & 68.1          \\
				Chen \etal~\cite{anatomypose}($T$=243)           & TCSVT2021      & 87.8          & 53.8          & 79.1          \\
				Gong \etal~\cite{PoseAug}                      & CVPR2021       & 88.6          & 57.3          & 73.0            \\
				Zheng \etal~\cite{poseformer}                   & ICCV2021       & 88.6          & 56.4          & 77.1          \\
				\rowcolor[HTML]{DADADA} 
				Ours($T$=1)               & \textbf{}      & \uline{94.2}          & \uline{63.8}          & \uline{57.9}          \\
				\rowcolor[HTML]{DADADA} 
				Ours($T$=27)              & \textbf{}      & \textbf{94.4} & \textbf{66.5} & \textbf{54.9}	\\
				\bottomrule
			\end{tabular}
		}
		\caption{Detailed quantitative comparison results on MPI-INF-3DHP with three metrics. The $\uparrow$ indicates the higher, the better, the $\downarrow$ indicates the lower, the better. The best and second-best results are highlighted in bold and underlined formats, respectively.}
		\label{tab:3dhp}
	\end{table}
	\begin{table}[htp]
		\centering
		\resizebox{230pt}{!}{
			\begin{tabular}{l|ccc|ccc|c}
				\toprule
				\textbf{\#Protocol1} & \multicolumn{3}{c}{Walk}                      & \multicolumn{3}{c}{Jog}                       & \textbf{Avg.}           \\
				\midrule
				Pavllo \etal~\cite{videopose3d}($T$=81)     & 13.1        & \textbf{10.1}            & 39.8          & 20.7            & 13.9          & 15.6          & 18.9          \\
				Pavllo \etal~\cite{videopose3d}($T$=81, FT) & 14.0 	& 12.5          & 27.1       & \textbf{20.3} & 17.9 & 17.5	& 18.2          \\
				Zheng \etal~\cite{poseformer}($T$=43)     & 16.3          & 11            & 47.1          & 25            & 15.2          & 15.1          & 21.6          \\
				Zheng \etal~\cite{poseformer}($T$=43, FT) & 14.4 & 10.2 & 46.6          & 22.7          & \textbf{13.4} & \textbf{13.4} & 20.1          \\
				\rowcolor[HTML]{DADADA}
				Ours($T$=43)           & 20.3          & 22.4          & 34.8          & 27.3          & 32.1          & 34.3          & 28.5          \\
				\rowcolor[HTML]{DADADA}
				Ours($T$=43, interval=1)           & 16.2          & 14.2          & 21.6          & 24.6          & 23.2          & 25.8          & 20.9          \\
				\rowcolor[HTML]{DADADA}
				Ours($T$=43, FT)       & \textbf{12.7}          & 10.9          & \textbf{17.6} & 22.6 & 15.8          & 17.0            & \textbf{16.1} \\
				\bottomrule
			\end{tabular}
		}
		\caption{The MPJPE on HumanEva testset under Protocol 1. FT indicates using the pretrained model on Human3.6M for finetuning. The best result is highlighted in bold.}
		\label{tab:heva_ft}
	\end{table}
	
	\subsection{Ablation Study}
	\label{sec:ablation}
	To evaluate the impact and performance of each component in our model, we evaluate their effectiveness in this section.
	The Human3.6M dataset and the CPN\cite{chen2018cascaded} detector are employed to provide 2D keypoints.
	
	\textbf{Effect of Each Component.}
	\label{sec:ablation_component}
	As shown in \Cref{tab:ablation_component}, we first modify the central frame 3D pose output to the sequence output without any other optimization to get the \textit{seq2seq} baseline model.
	For a fair comparison, the parameter setting of the \textit{seq2seq} baseline is directly applied to the proposed method, and the MPJPE loss is utilized in the baseline model.
	After applying the alternating design, the result shows that our method decreases 6.2mm MPJPE (from 51.7mm to 45.5mm).
	Then joint separation is utilized to demonstrate its advantage in both improving the performance (from 45.5 to 41.7) and reducing computing cost (FLOPs for each frame decreases to 645 from 186405). 
	By applying our loss function to replace MPJPE loss, our result achieves the best (40.9mm MPJPE with 645 FLOPs).
	The MixSTE with our loss function improves 20.9\% (from 51.7 to 40.9) compared to the \textit{seq2seq} baseline, and it proves the rationality of our network design.
	\begin{table}[bp]
		\centering
		\resizebox{230pt}{!}{%
			\begin{tabular}{lcccc|cc}
				\toprule
				& Seq2seq & \makecell[c]{Alternating \\Design} & \makecell[c]{Joint \\Separation} & \makecell[c]{Our\\Loss} & MPJPE & FLOPs (M) \\
				\midrule
				Baseline & \Checkmark  &   &   &  & 51.7	& 186405\\
				& \Checkmark  & \Checkmark   &    &            & 45.5  & 186405\\
				& \Checkmark       & \Checkmark              & \Checkmark       &    & 41.7  & 645\\
				\rowcolor[HTML]{DADADA}
				Ours     & \Checkmark & \Checkmark  & \Checkmark   & \Checkmark		& 40.9  & 645\\
				\bottomrule  
			\end{tabular}%
		}
		\caption{Ablation study for each component used in our method. The evaluation is performed on Human3.6M with MPJPE (mm) and FLOPs.}
		\label{tab:ablation_component}
		\vspace{-0.3cm}
	\end{table}
	
	\begin{table}[bp]
		\centering
		\resizebox{190pt}{!}{
			\begin{tabular}{l|c|c}
				\toprule
				& MPJPE & MPJVE \\
				\midrule
				MPJPE Loss                         & 41.7	& 5.0	\\
				WMPJPE Loss                        & 41.3	& 4.6	\\
				WMPJPE Loss + Motion Loss~\cite{motionguidepose}		   & 41.3	& 4.3	\\
				WMPJPE Loss + TCLoss~\cite{temporalconsis}		& 41.2 	& 3.6	\\		
				WMPJPE Loss + MPJVE Loss		   & 41.2	& 2.6	\\
				\rowcolor[HTML]{DADADA}
				Ours (WMPJPE Loss + T-Loss) 	   & 40.9	& 2.3	\\
				\bottomrule
			\end{tabular}
		}
		\caption{Ablation study for loss function in our method with MPJPE and MPJVE.}
		\label{tab:ablation_loss}
		\vspace{-0.6cm}
	\end{table}
	
	\textbf{Effect of Loss Function. }	
	We have explored the contribution of our loss function in detail.
	As shown in \Cref{tab:ablation_loss}, the MPJPE metric decreases from 41.7 to 41.3 after applying the WMPJPE loss.
	The result demonstrates that the WMPJPE is an essential loss to improve accuracy. 
	Then the temporal consistency loss (TCLoss) following~\cite{temporalconsis} is employed to improve the temporal smoothness performance (MPJVE) by 1.0 (decreases from 4.6 to 3.6), and the coherence gets better after using the MPJVE loss (decreases from 4.6 to 2.6).
	The motion loss~\cite{motionguidepose} has less contribution to the coherence than TCLoss and MPJVE loss.
	Finally, after applying the T-Loss and WMPJPE loss to our method, the result achieves the best on the MPJPE and MPJVE metrics (40.9mm MPJPE, 2.3 MPJVE).
	The ablation study demonstrates that our loss function is comprehensive for the proposed model regarding accuracy and smoothness.
	
	
	\textbf{Parameter Setting Analysis.}
	\Cref{tab:ablation_params} shows how the setting of different hyper-parameters in our method impacts the performance under Protocol 1 with MPJPE. 
	There are three main hyper-parameters for the network: the depth of MixSTE ($d_l$), the dimension of model ($d_m$), and the input sequence length ($T$).
	We divide the configurations into 3 groups row-wise, and different values are assigned for one hyper-parameters while keeping the other two hyper-parameters fixed to evaluate the impact and choice of each configuration.
	Based on the results in the table, we choose the combination of $Depth$=8, $Channel$=512, and $Input\ Length$=243.
	Note that we choose the $Depth=8$ rather than $Depth=10$ because the latter setting introduces a more significant number of parameters (33.7M \vs 42.2M).
	\begin{table}[hp]
		\centering
		\resizebox{190pt}{!}{
			\begin{tabular}{ccc|c}
				\toprule
				Depth ($d_l$) & Dimension ($d_m$)    & Input Length ($T$) & MPJPE \\
				\midrule
				4      & 64            & 27    & 54.3      \\
				6      & 64            & 27    & 53.2      \\
				8      & 64            & 27    & 51.8      \\
				\textbf{10}     & 64            & 27    & 51.1      \\
				\hline
				8      & 128           & 27    & 47.9      \\
				8      & 256           & 27    & 46.1      \\
				8      & \textbf{512}           & 27    & 45.1      \\
				8      & 640 		   & 27    & 46.0      \\
				\hline
				8      & 512           & 81    & 42.7      \\
				8      & 512           & 128   & 42.0      \\
				\rowcolor[HTML]{DADADA}
				8      & 512           & \textbf{243}   & \textbf{40.9}       \\
				8      & 512           & 300   & 41.8      \\
				\bottomrule
			\end{tabular}
		}
		\caption{Ablation study for hyper-parameter setting in depth ($d_l$), dimension ($d_m$) and input length ($T$). The evaluation is performed on Human3.6M with MPJPE (mm).}
		\label{tab:ablation_params}
	\end{table}
	
	\subsection{Qualitative Results}
\begin{figure}[htp]
		\centering
		\includegraphics[width=1.0\linewidth]{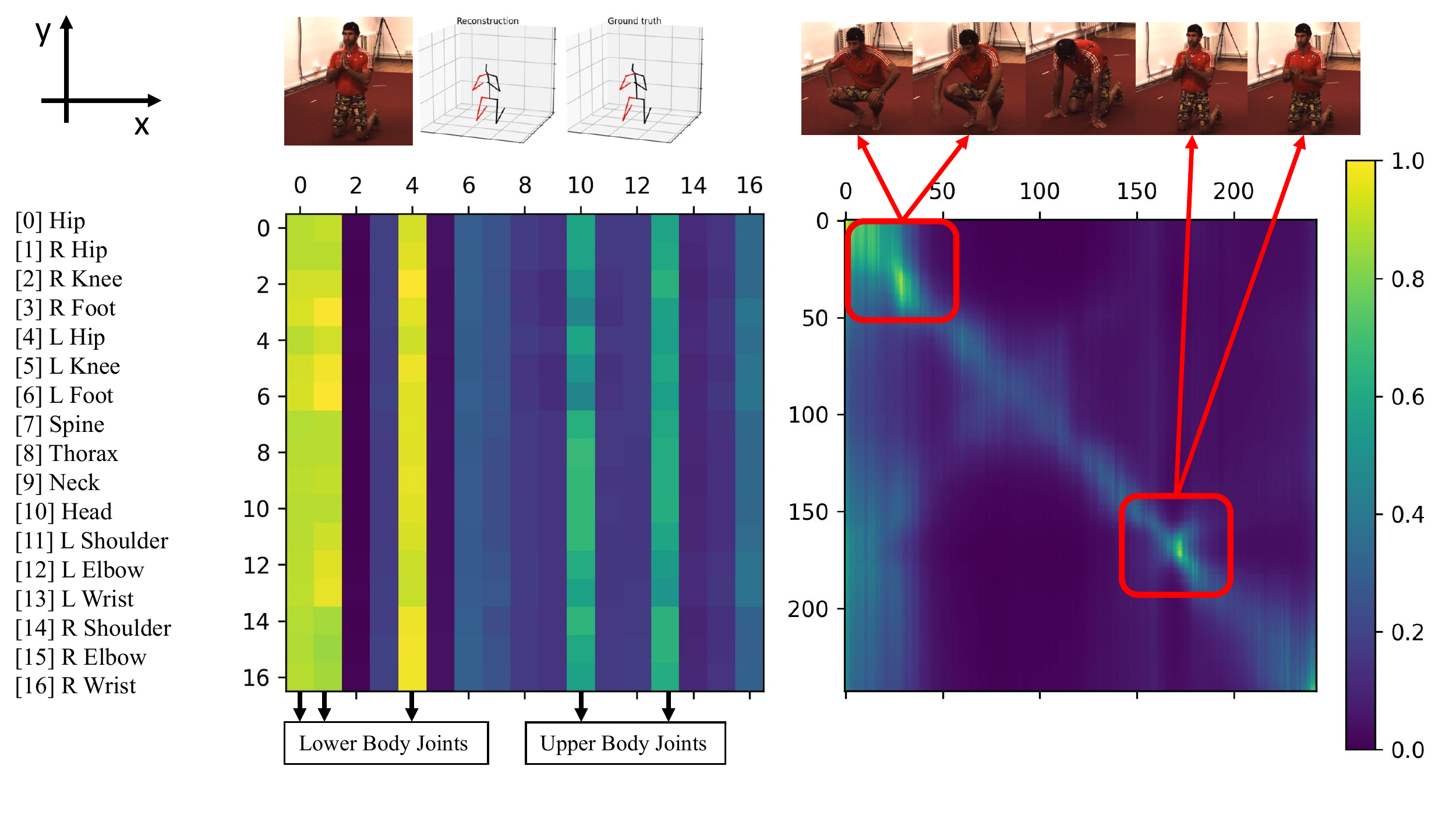}
		\vspace{-0.9cm}
		\caption{
			Visualization of self-attentions among body joints and frames. 
			The x-axis and y-axis correspond to the queries and the predicted outputs, respectively. 
			Each row shows the attention weight $w_{i,j}$ of the $j$-th query for the $i$-th output.
		}
		\label{fig:attnmap}
		\vspace{-0.3cm}
	\end{figure}

\begin{figure}[htp]
		\centering
		\includegraphics[width=0.9\linewidth]{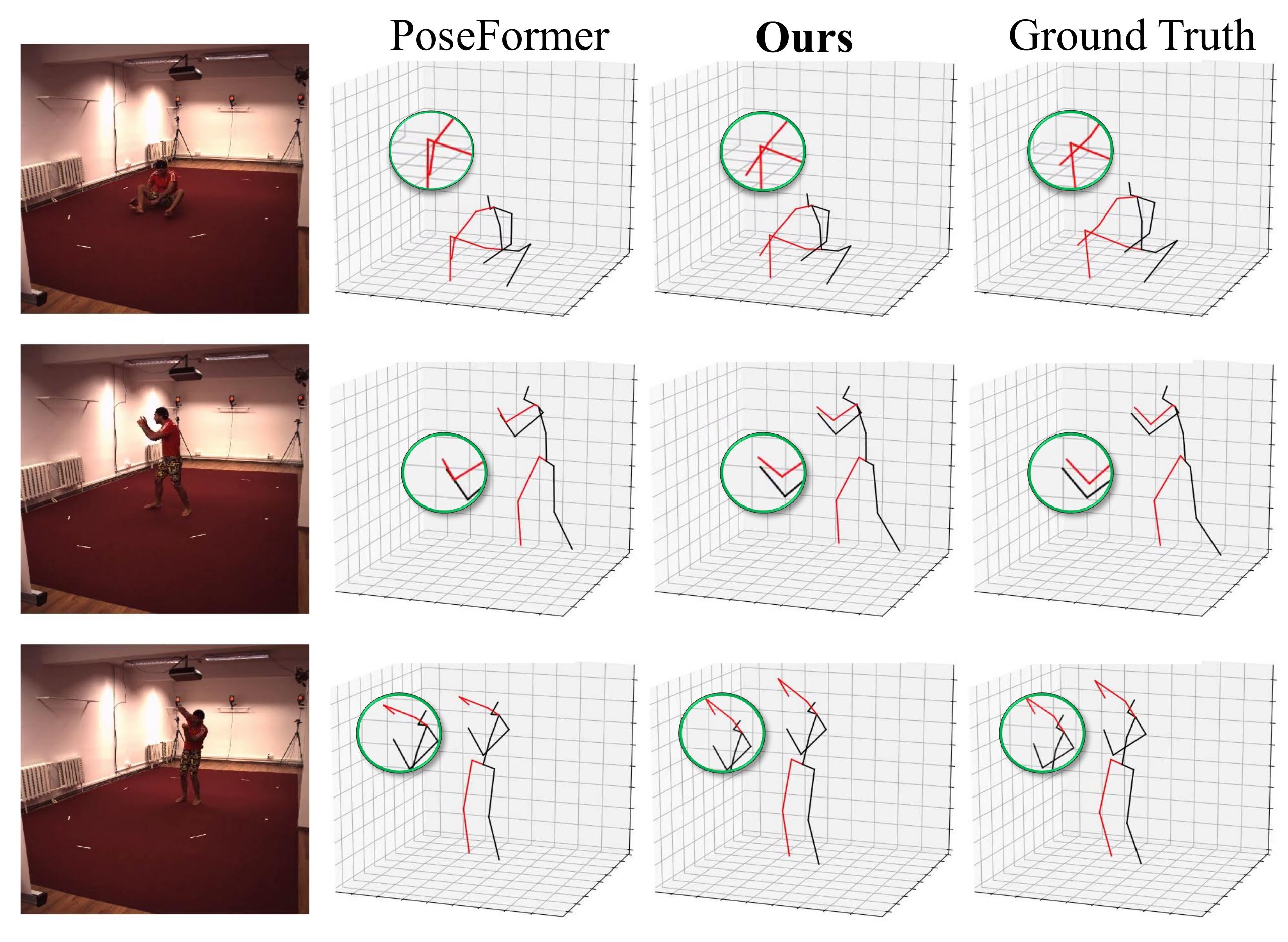}
		\caption{Qualitative comparison between our method (MixSTE) and~\cite{poseformer} with the \textit{Photo} and \textit{SittingDown} actions on on Human3.6M. The green circle highlights locations where our method has better results.}
		\label{fig:demo}
		\vspace{-0.3cm}
	\end{figure}

	As shown in \Cref{fig:attnmap}, we further conduct visualization on spatial and temporal attention. 
	The selected action (\textit{SittingDown} of testset \textit{S11}) is applied for visualization. 
	Moreover, attention outputs of different heads are averaged to observe the overall correlations of joints and frames, and the attention outputs are normalized to $[0,1]$. 
	It can be easily observed from spatial attention map (left of \Cref{fig:attnmap}) that our model learns different dependencies between joints.
	Furthermore, we also visualize the temporal attention map (right of \Cref{fig:attnmap}) from the last temporal attention layer. 
	The two parts with light color have similar poses with adjacent frames, while the dark color corresponded frame (the middle image in the frame sequence) has a more different pose with adjacent frames.
	We also evaluate the visual result of estimated poses and 3D ground truth of Human3.6M in \Cref{fig:demo} to show that we can estimate more accurate poses compared to PoseFormer~\cite{poseformer}.
	
	\section{Conclusion}
	We have presented MixSTE, a novel transformer-based \textit{seq2seq} approach for 3D pose estimation from monocular video.
	The model can better capture global sequence coherence and temporal motion trajectories of different body joints.
	Moreover, the efficiency of 3D human pose estimation is much improved.
	Comprehensive evaluation results show that our model obtains the best performance.
	As a new universal baseline, the proposed method also opens up many possible directions for future works.
	Nonethless, our method is still limited by inaccurate 2D detection results \eg missing and noisy keypoints.
	It may be alleviated by applying better 2D detector, but modeling distribution of input noise is also a feasible and valuable exploration.\\
	\textbf{Acknowledgements.} This work was supported by the National Natural Science Foundation of China under Grant 62106177 and 61773272.

	{\small
		\bibliographystyle{ieee_fullname}
		\bibliography{Mypaper.bib}
	}
	
\end{document}